\documentclass[letterpaper,10pt]{article}

\usepackage[preprint]{probml}

\ShortHeadings{Uncertainty-aware glucose prediction}

\usepackage{blindtext}
\usepackage{wrapfig}
\usepackage{enumitem}
\usepackage{graphicx}
\usepackage[table]{xcolor}
\usepackage{subcaption}
\usepackage{caption}
\usepackage[font=footnotesize]{caption} 

\usepackage{bm,amssymb,amsmath}

\makeatletter
\def\th@plain{%
  \thm@notefont{}
  \itshape 
}
\def\th@definition{%
  \thm@notefont{}
  \normalfont 
}
\makeatother

\def\1{\bm{1}}

\DeclareMathAlphabet{\mathsfit}{\encodingdefault}{\sfdefault}{m}{sl}
\SetMathAlphabet{\mathsfit}{bold}{\encodingdefault}{\sfdefault}{bx}{n}

\def\be{\begin{equation}}
\def\ee{\end{equation}}
\def\bea{\begin{eqnarray}}
\def\eea{\end{eqnarray}}

\defcitealias{ADA}{ADAPP 2024}

\begin{document}

\title{
Uncertainty quantification in neural network-based glucose prediction for diabetes
}

\author[1,$*$,$\dagger$]{Hai Siong Tan}
\author[2,$*$]{Rafe McBeth}

\affil[1]{Gryphon Center for A.I. and Theoretical Sciences, Singapore}
\affil[2]{University of Pennsylvania, Perelman School of Medicine, Philadelphia, USA}
\affil[$*$]{Equal contribution}
\affil[$\dagger$]{Correspondence to \url{haisiong.tan@gryphonai.com.sg}}

\maketitle

\begin{abstract}
In this work, we investigate uncertainty-aware neural network models for blood glucose prediction and adverse glycemic event identification in Type 1 diabetes. We consider three families of sequence models based on LSTM, GRU, and Transformer architectures, with uncertainty quantification enabled by either Monte Carlo dropout or through evidential output layers compatible with Deep Evidential Regression. Using the HUPA-UCM diabetes dataset for validation, we find that Transformer-based models equipped with evidential output heads provide the most effective uncertainty-aware framework, achieving consistently higher predictive accuracies and better-calibrated uncertainty estimates whose magnitudes significantly correlate with prediction errors. We further evaluate the clinical risk of each model using the recently proposed Diabetes Technology Society error grid, with risk categories defined by international expert consensus. Our results demonstrate the value of integrating principled uncertainty quantification into real-time machine-learning-based blood glucose prediction systems.
\end{abstract}

\section{Introduction}
\label{sec:intro}

This work explores machine learning algorithms that generate 
blood glucose prediction for diabetic patients based on past records of glucose and bolus insulin tracked at regular intervals by Continuous glucose monitoring (CGM) systems, and other physiology-related data. As emphasized by a global consensus of diabetes clinicians and technologists in \citet{consensus_2022,cgm,consensus_2017}, CGM enables real-life tracking of blood glucose levels at a fixed sampling rate and has been established as a powerful method in diabetes management \citep{glimpse}. It can provide time series data suitable for training machine learning models \citep{review2,review} in glucose prediction tasks. Having an AI-guided glucose forecasting system supports diabetes management in at least the following two areas \citep{Zhu,review,diabits}: (i)timely adjustments of diet, exercise and insulin doses to avoid potential hypoglycemia (glucose level $<70$ mg/dl) and hyperglycemia (glucose level $>180$ mg/dl), both adverse glycemic events being of primary interest to clinical diabetes management, 
(ii)supporting a closed-loop system involving a continuous insulin pump for diabetic patients. While hyperglycemia is associated with short-term complications like ketoacidosis, acute hypoglycemia can lead to severe convulsions and coma \citep{ADA,medical}, apart from being a major impediment to diabetes therapy intensification \citep{rama}.
Against the backdrop of potentially severe health consequences in adverse glycemic events, we would like our machine learning models to be reliable in the sense of being equipped with robust uncertainty quantification.

Thus, in contrast to the majority of past works in this area \citep{review}, here
our focus lies in examining the extent to which uncertainty quantification aids in enhancing the effectiveness of the glucose prediction system, the main uncertainty quantification methodologies we examine being Monte Carlo dropout \citep{dropout} and Deep Evidential Regression \citep{amini} techniques. These are incorporated into three families of models based on (i)Long Short-Term Memory (LSTM) \citep{lstm} (ii)Gated Recurrent Units (GRU) \citep{gru} and (iii)Transformer \citep{transf} architectures respectively. 
We validate our models on the publicly accessible HUPA-UCM dataset of \citet{HUPA} which consists of real clinical data from a medical study of 25 patients with Type 1 diabetes and leverages measurements from CGM devices. Like related recent works in \citet{Esha,Zhu}, apart from examining prediction accuracy at some fixed-duration horizon, we also assess our models' effectiveness at predicting occurrences of hypoglycemia and hyperglycemia. This classification task is of particular importance for Type 1 diabetes patients who require injection of exogenous artificial insulin in the absence of natural insulin secretion \citep{ADA}. Developing effective uncertainty-aware glucose prediction models that leverage CGM data is a natural goal towards realizing AI technology that can be used in a reliable way to manage diabetes conditions on a daily basis.


\section{Related Work}

We refer the reader to \citet{review2,review3,review} for recent reviews of glucose prediction models in literature, yet we would like to briefly comment on two particular papers which are more intimately connected to our approach. 
To our knowledge,  
the model proposed in \citet{Zhu} was the only prior work in published literature on glucose prediction that examined an uncertainty distribution-equipped model. They had also invoked evidential regression in constructing their model
named as `Fast-adaptive and Confident Neural Network', the basic architecture being a stack of 3 bi-directional GRU layers followed by an attention layer preceding the evidential outputs corresponding to the parameters of the t-distribution. 
Model training was performed on the OhioT1DM dataset \citep{Ohio} and two other proprietary datasets (of Imperial College London, UK) with roughly a 3:1:1 training/validation/testing chronological split. The loss function is the loss term in eqn.~\eqref{dataloss} with the regularization term proposed by Amini et al. in \citet{amini}. The input sequences were (i)CGM glucose (ii)bolus insulin (iii)carbohydrate intake. Similar to our setup, prediction horizons were taken to be 30 and 60 minutes. However, there were some aspects of uncertainty quantification that were significantly different from our approach: (i)the confidence bands were constructed by taking an empirical parameter representing a `z-value' to the epistemic uncertainty expressed in eqn.~\eqref{uncertainties}. 
In our understanding, this is not quite the correct confidence interval as the predictive variance of the t-distribution is the sum of the aleatoric and epistemic uncertainties, (ii)there was no usage of any metric to assess calibration of uncertainty estimates in \citet{Zhu}, (iii)there was no comparison to other uncertainty quantification methods. 

In \citet{Esha}, the authors compared LSTM models trained on individual-specific data to those trained on population-level data from the HUPA-UCM dataset. They found that individualized models achieved comparable prediction accuracy to aggregated models and Clarke error grid Zone A accuracies. Subject-level analyses revealed only modest differences between the two approaches, with some individuals benefiting more from personalized training. The focus in \citet{Esha} was in examining whether training models on 
individual-based or population-based data leads to significant differences in results rather than any usage/study of uncertainty quantification techniques. Thus, similar to \citet{Zhu} and other previous works covered in the reviews \citet{review2,review}, there was no usage of uncertainty quantification-related metrics. Their choice of input features was one of the four classes that we have used in our work here: \{ glucose, bolus and basal insulin, carbohydrate intake \}.

\section{Methodology}
\subsection{On the HUPA-UCM dataset}

This is a dataset \citep{HUPA}
furnishing real data
from a medical study of 25 patients with Type 1 diabetes mellitus (T1DM). It provides CGM data, insulin dose (both bolus volumes and a continuous basal insulin rate) and
related physiological quantities: heart rate, steps, calories burned, meal ingestion estimated in carbohydrate grams.
CGM data was acquired by FreeStyle Libre 2 systems, while Fitbit Ionic smart watches were used for measuring steps, calories and heart rate data. The sampling rate was five minutes and data for each patient was collected for at least 2 weeks under free-living conditions. This HUPA-UCM dataset was curated by researchers at the \emph{Complutense University of Madrid} and Prince of Asturias University Hospital, and made publicly accessible for research purposes without any additional approval procedures required. In our work here, we used the provided preprocessed version of the dataset which required no additional cleaning (e.g. interpolation for missing values, etc.).

\subsection{Aspects of model implementation}

For each patient, data were split chronologically into 60\% training, 20\% validation, and 20\% testing to avoid temporal leakage. Inputs consist of 4 feature sequences : blood glucose, bolus insulin, carbohydrates intake and one of the following $\{\text{heart rate}, \text{steps}, \text{calories}, \text{basal insulin}\}$, with all features standardized using z-score normalization. We focus on the prediction horizons $h_L =$ 30 min and 1 hour. Models were trained for 300 epochs using Adam (learning rate $10^{-4}$, batch size 1024). Dropout-based models were optimized with mean squared error, whereas evidential models used 
$\mathcal{L}_{evidential} = \mathcal{L}_{data} + 0.01*\mathcal{L}_{kl}$
where $\mathcal{L}_{data}$ is defined in eqn.~\eqref{dataloss} and $\mathcal{L}_{kl}$ defined in eqn.~\eqref{kl}. 
Validation data were used for each model's hyperparameter selection, and the test set was reserved for final evaluation. All models were trained on an NVIDIA A100 GPU. In the following, we briefly summarize the three neural network base models considered in our experiments and the classical ridge regression baseline method \citep{ridge}.

\begin{enumerate}
\item{\textbf{LSTM:}}
For this class of models, we used unidirectional LSTMs that map the four-dimensional input sequences to glucose trajectories. The sequence was processed by a single-layer LSTM with hidden size 128, and the hidden state of the final timestep was used as a compact representation of the historical context. This representation was passed through a fully connected prediction head consisting of a 64-unit linear layer followed by a dropout layer (activated for the model equipped with uncertainty quantification) and a final linear output layer that produced the 
$h_L$-step glucose forecast.

\item{\textbf{GRU:}} We considered an attentive bidirectional GRU architecture where the input sequence of four-dimensional input sequence was processed by a 3-layer bidirectional GRU with hidden size of 40 per direction, producing temporal representations of dimension 80 at each timestep. A temporal attention mechanism was then applied to the full sequence to compute a weighted context vector summarizing the history, allowing the model to focus on clinically relevant past observations. The resulting context representation was passed through a fully connected prediction head consisting of a 32-unit hidden layer with ReLU activation and a dropout layer, followed by a linear output layer that produced the 
$h_L$-step glucose forecast.

\item{\textbf{Transformers:}} We used a causal Transformer encoder where the four-dimensional input sequence was first projected to a model dimension of 64 and augmented with positional encoding. The sequence was then processed by a 2-layer Transformer encoder composed of multi-head self-attention with 4 attention heads and feedforward sub-layers of dimension 128 using GELU activation. A causal attention mask ensured that each timestep attended only to past observations. The hidden representation of the final timestep was used as the sequence summary and passed through a linear output layer to produce the $h_L$-step glucose forecast.

\item{\textbf{Ridge Regression:}} As a non-neural network classical baseline, we used Bayesian ridge regression \citep{ridge}. A Gaussian prior was imposed on the regression weights, and the posterior was computed analytically. Predictions for new input sequences were assumed to be Gaussian distributed, with mean equal to the posterior mean prediction and variance accounting for both residual noise and parameter uncertainty.

\end{enumerate}

Monte Carlo dropout leverages the dropout layers introduced in each of the base model architecture by keeping them active at test time and performing multiple stochastic forward passes \citep{dropout}. This process approximates sampling from a posterior distribution over the network weights, with predictive uncertainty quantified from the variance across the resulting outputs. For evidential regression-based uncertainty quantification, we expand the output layer of each base model such that for each forecast horizon step, the model outputs the four parameters of the evidential predictive distribution which can be used to compute uncertainty estimates. In Appendix \ref{apd:evidential}, we provide the technical details of evidential regression including a derivation of the loss function. For Monte Carlo dropout-based models, we took the mean-squared-error as the loss function.

\subsection{On evaluation of model accuracy}

Related literature on blood glucose prediction from CGM datasets often used the Clarke Error Grid \citep{clarke} as one of the metrics for prediction models. Designed specifically for clinical decision-making, an error grid generally describes the clinical point accuracy of a glucose monitor compared with reference values, with reference/monitor paired values assigned to clinical risk zones. Using such an error grid to assess accuracy of machine-learning-based prediction models enables one to evaluate them in a manner that is more informative with regards to the clinical risks they carry. In our work, we used a recent error grid proposed by an international consensus of diabetes clinicians that is designed to replace the Clarke Error Grid - the Diabetes Technology Society (DTS) Error Grid in \cite{DTS}.
There are five clinical risk zones (A-E) representing 
$\{$ `no', `mild', `moderate', `high', `extreme' $\}$ risks respectively. 
For model evaluation, we focused on the percentage of predictions falling within zone A (no clinical risk) as a representative indicator of the model's clinical reliability. Mathematically, this is a thin region surrounding the diagonal in the prediction-ground truth plane that clinicians define as practically the set of glucose values for which model accuracy is enough for safe clinical decision-making\footnote{See Supplementary materials of \citet{DTS} for the precise geometric definitions of each zone.}. Another crucial metric of accuracy that we used is that of 
the mean absolute relative difference (MARD). 
As described in \citet{consensus_2022,consensus_2017},
MARD is the most established statistic for assessing the accuracy of CGM devices with respect to corresponding values of some reference system. Here we used it as a measure of model accuracy with respect to the measurements of the CGM device. 

\subsection{Metrics for assessing uncertainty quantification}

To assess uncertainty quantification, we used the following metrics: (i)Brier scores, (ii)Precision-Recall AUC, (iii)Spearman's correlation between uncertainty estimates and prediction errors ($\equiv \rho$),
(iv)Spearman's correlation between uncertainty estimates and the DTS clinical risk zones ($\equiv \rho_z$),
and (v)reliability diagrams plotting empirical coverage probabilities (ECP) against nominal ones, and the mean calibration error (MCE) defined as the mean absolute value of the deviation between ECP and nominal probabilities.  

The Brier score \citep{brier} is defined as the average 
$
\frac{1}{N}
\sum^N_{i=1} (g_i - y_i)^2,
$
where $g_i, y_i$ are the model probabilities and observed occurrences for these events respectively, and $N$ refers to the total number of testing data points. 
This proper scoring rule evaluates the entire predicted probability distribution. For each model, we evaluated the sensitivity and Precision-Recall AUC (PR-AUC) for hypo- and hyperglycemia detection. 
We also relied on the Spearman's coefficients $\rho, \rho_z$ to quantify the degree to which the uncertainty estimates correlated with predictive errors and clinical risks as defined by the consensus of human experts in \cite{DTS}.
We represented the uncertainty estimates by the standard deviation of the uncertainty's probability distribution. In the case of evidential regression-based models, this was $\sigma_p$ as defined in eqn.~\ref{uncertainties} of Appendix \ref{apd:evidential}. For dropout models, this was the empirical standard deviation of the Monte-Carlo samples, whereas for ridge regression, we used the known analytical formula for the standard deviation of the assumed underlying Gaussian distribution \citep{ridge}. In particular, $\rho_z$ connects the model’s uncertainty quantification to the diabetes physicians' knowledge encoded in the risk stratification scheme of \citet{DTS}. Non-trivial correlations would then suggest that the model’s uncertainty estimates reflect clinically relevant notions of reliability. To compute $\rho_z$, we represented each DTS clinical risk zone as an ordinal integer from 1 - 5 pertaining to risk zones A - E respectively. $\rho_z$ was then the ordinary Spearman's coefficient between the uncertainty estimates and the DTS clinical risk zone associated with each model prediction. 

Another metric that we used to assess uncertainty quantification was through the empirical coverage probability. For each nominal coverage level $1 - \alpha$, we computed the two-sided predictive intervals using the learned t-distribution in the case of evidential models or empirical quantiles of the samples for dropout-based model.
The empirical coverage probability was simply defined as the fraction of observed glucose values contained within the interval. Calibration was assessed by plotting empirical versus nominal coverage and summarized by the mean absolute coverage error, i.e., the average absolute deviation between empirical and nominal coverage across confidence levels. These metrics have traditionally been used for uncertainty quantification in other contexts of machine learning \citep{ECP,jungo}. 

\section{Results}

 In the following, model performance is analyzed in the aspects of accuracy, the degree of calibration of the uncertainty quantification, and through several forecast diagrams which illustrate how uncertainty estimates can be useful in glucose prediction. Table~\ref{tab:heart_short} gathers the evaluation metrics for models trained on heart rate-included inputs for a 30-min predictive horizon. It acts as a performance summary for our work
 as it contains trends that are shared across models trained on other input sets as tabulated in Tables~\ref{tab:basal_short} - \ref{tab:heart_long} in Appendix~\ref{apd:other_tables}.

\subsection{On prediction and adverse glycemic event classification accuracy}

In terms of the DTS Error Grid zone A accuracy, a specific model clearly stood out -- Transformer-based model equipped with evidential regression output layers, hereafter referred to as Transformer Evidential Model (TEM). It achieved the best performance across all experiments. For each experiment, all neural network models scored higher zone A accuracies than the ridge regression baseline. In the aspect of MARD, TEM attained the lowest value consistently across all experiments and again, whereas ridge regression was the worst model in each case. The mean degradation of zone A accuracy across all considered models with 1-hr predictive horizon relative to the 30-min case was about $8 \%$, accompanied by an average increase in MARD of about $4 \%$. In the aspect of sensitivity in detecting hypoglycemia, we found that evidential regression-equipped models and ridge regression yielded the highest values for both horizon durations, with a sensitivity of $\sim 0.9$ for the 1-hr horizon and about $\sim 5\%$ higher for the 30-min case. For hyperglycemia, evidential and ridge regression-based models similarly scored higher sensitivities.

At the level of per-patient metrics, 
we also investigated the sensitivity of our accuracy indices : the DTS zone A accuracy and MARD on the choice of input features, as we would like to know whether our results would reveal any particular choice of input features to be more discriminatory. Visual inspection of the population-level results in Tables~\ref{tab:heart_short} to \ref{tab:heart_long} appears to show no significant dependence. 
We performed Friedman tests on the per-patient MARD and zone A 
accuracy scores attained by TEM across all four classes of inputs and for both horizon lengths. For MARD values in the 30-min horizon setting, Friedman test revealed a statistically significant effect of input feature choice ($\chi^2(3) = 11.9,\,p < 0.01)$, with heart rate-trained models being the best based on mean ranks. Effect size estimated using Kendall's W was 0.16, indicating a small but consistent difference in ranking across patients. Post-hoc Wilcoxon signed-rank tests with Holm correction indicated heart rate-trained models to be better in MARD relative to those trained on steps and calories in the short horizon setting (Holm-adjusted $p \sim 0.01$) relative to that trained on steps in the long horizon setting (Holm-adjusted $p \sim 0.04$). For both types of accuracy metrics and horizon durations, heart rate-trained models yielded the best mean scores. Based on these results, it thus appears that there is a slight advantage in model accuracy for heart rate-included inputs compared to the other three input classes defined by the inclusion of steps, calories and basal insulin.\footnote{In the Appendix, Figures~\ref{fig:mard_feature} and \ref{fig:zoneA_feature} show the boxplots for visualizing input feature dependence of per-patient accuracy metrics.}

\begin{table*}
\centering
  \caption{Models trained on heart rate-included inputs (30 min horizon). Abbreviations: 
zA - zone A accuracy; 
MARD - mean absolute relative difference;
${}^{70}S$ - hypoglycemia sensitivity; ${}^{70}B$ - hypoglycemia Brier score; 
${}^{70}A$ - hypoglycemia PR-AUC; ${}^{180}\{ S, B, A\}$ - the corresponding metrics for hyperglycemia; MCE - mean calibration error; $\rho$ - Spearman's correlation between ucertainty and error; model subscripts {\itshape d, e} - equipped with Monte Carlo dropout and evidential regression respectively; $\rho_z$ - Spearman's correlation between uncertainty and clinical risk zones as defined in \cite{DTS}. Each shaded cell pertains to the best-performing model for each metric column.
}
\vspace{5pt}
\label{tab:heart_short}
  {\small
  \begin{tabular}{c | c c | c c c |c c c| c c c }
    \toprule
    Model & zA & MARD & ${}^{70}S$ & ${}^{70}B$ &  ${}^{70}A$ &
    ${}^{180}S$ & 
    ${}^{180}B$ &  ${}^{180}A$ & MCE & $\rho$ & $\rho_z$\\
    \hline
    LSTM & 96.7 & 4.52 & 0.80  & ${}$ &  0.91 & & &
    \\
    $\text{LSTM}_d$ & 96.6 & 4.66 & 0.88 & 0.018 & 0.89
    & 0.92  & 0.023 & 0.95 &
    0.10 & 0.15 & 0.073
    \\
    $\text{LSTM}_e$ & 96.6 & 4.24 & \cellcolor{gray!20} 0.94 & 0.016 & 0.93 & \cellcolor{gray!20} 0.96 & 0.020 & 0.98 & 0.04 & 0.67 & 0.201
    \\ 
    $\text{Transf}$ & 96.7 & 4.44  & 0.78 & ${}$ & ${}$ & 0.91  & ${}$ & ${}$ & ${}$ 
    \\
    $\text{Transf}_d$ & \cellcolor{gray!20} 96.8 & 4.50 & 0.87 & 0.018 & 0.90 & 0.93 & 0.022 & 0.95 & 0.09 & 0.23 & 0.098
    \\
    $\text{Transf}_e$ & \cellcolor{gray!20} 96.8 & \cellcolor{gray!20} 4.14 & \cellcolor{gray!20}0.94  & \cellcolor{gray!20} 0.015 & \cellcolor{gray!20}0.94 & \cellcolor{gray!20} 0.96  & \cellcolor{gray!20}0.019 & \cellcolor{gray!20}0.99 & \cellcolor{gray!20}0.03 & \cellcolor{gray!20}0.68 
    &
    \cellcolor{gray!20}0.208
    \\
     $\text{GRU}$ & 96.7 & 4.49  &
     0.77 & & & 0.91  & ${}$ & ${}$ 
    \\
    $\text{GRU}_d$ & \cellcolor{gray!20} 96.8 & 4.80 & 0.91 & 0.020 & 0.90 & 0.93  & 0.022 & 0.96 & 0.07 & 0.19 & 0.065
    \\
    $\text{GRU}_e$ & 96.5 & 4.33 & \cellcolor{gray!20}0.94  & 0.016 & 0.92 &\cellcolor{gray!20} 0.96 & 0.021 & 0.98 & 0.05 & 0.63 & 0.176
    \\
    Ridge Reg. & 95.5 & 5.69  &\cellcolor{gray!20} 0.94 & 0.021 & 0.88 & \cellcolor{gray!20} 0.96 & 0.026 &0.97 & 0.12 & 0.48 & 0.203
    \\
    \bottomrule
  \end{tabular}}
\end{table*}

\subsection{On the degree of calibration of uncertainty estimates}

 We found that TEM models were associated with the highest Spearman's $\rho$ of $\sim 0.7$, with all evidential models yielding $\rho >0.6$, generally at least twice higher than the corresponding dropout-based models. Ridge regression models generally had $\rho \sim 0.5$. These results were consistently similar for both horizon periods and all the four classes of input sequences. In the aspect of $\rho_z$, evidential models tended to have relatively higher scores of about $\sim 0.3$ for the long horizon and $\sim 0.2$ for the short horizon, indicating a non-trivial positive correlation between uncertainty estimates and the discrete DTS clinical risk zones. Ridge regression also performed comparably in this index with slightly lower values whereas dropout models yielded $\rho_z \sim 0.1$. Among all models, TEM models performed the best with an average value of $\rho_z \sim 0.34$, showing that its uncertainty estimates were informative with respect to clinically defined risk zones.

\begin{figure}[ht]
 \centering
\includegraphics[width=\textwidth]{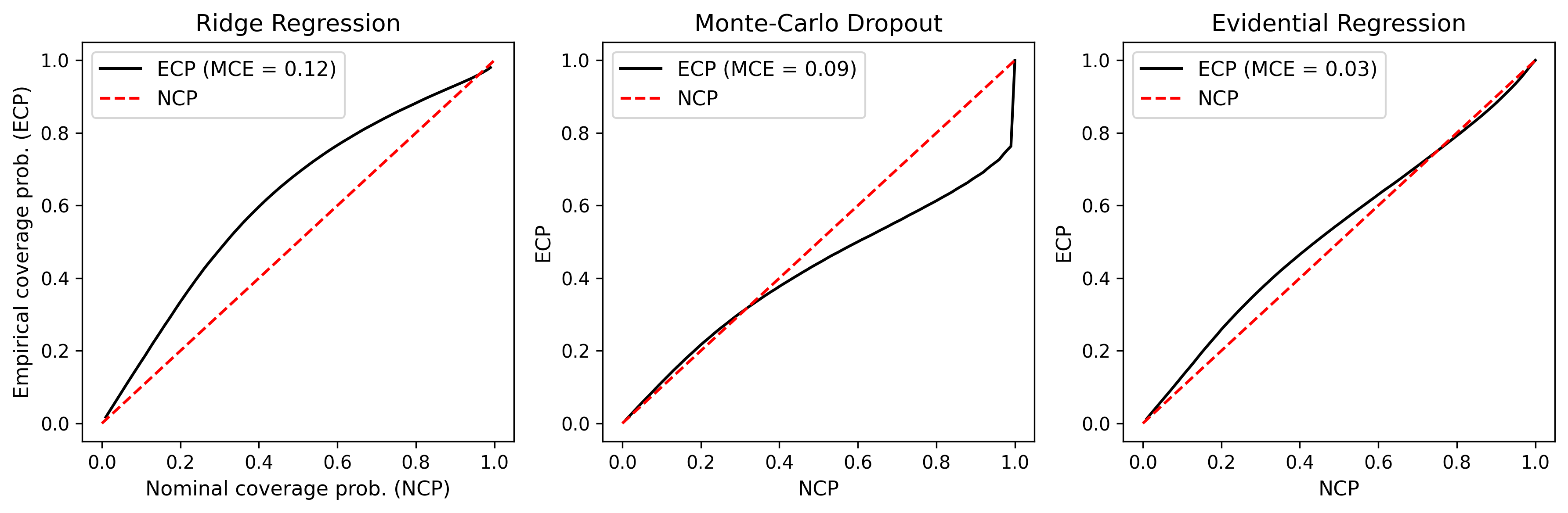}
  \caption{Plot of empirical coverage probabilities (ECP) vs nominal coverage probabilities (NCP) for three types of models trained on heart rate-included inputs and defined by a 30 min prediction horizon. The ideal plot is characterized by the ECP curve being very close to the nominal straight line joining the origin to the point (1,1). MCE denotes the mean calibration error which is the average of the absolute value of deviation between ECP and NCP. }
   \label{fig:ecp}
\end{figure}

In the aspect of MCE, evidential models yielded the lowest values compared to dropout-based models and ridge regression. In particular, TEM models attained the best scores in each experiment. In Fig.~\ref{fig:ecp}, we plot the ECP vs NCP plots for three different models trained on the heart rate-included inputs. As shown in Fig.~\ref{fig:ecp}, ridge regression's calibration plot reveals an under-confident uncertainty distribution whereas 
Monte Carlo dropout yielded an over-confident uncertainty distribution. When the same base model is equipped with an evidential output layer instead, the calibration plot shows an ECP curve that is very close to the ideal straight line, revealing a well-calibrated uncertainty quantification. In the aspect of Brier and PR-AUC scores, evidential models yielded the lowest and highest values respectively for both hypoglycemia and hyperglycemia detection, with the TEM models attaining the optimal scores among all models. Unlike the metric MCE which was horizon-independent, there was degradation of Brier and PR-AUC scores for the experiments defined with the 1-hr horizon relative to the 30-min case.

\subsection{Forecast Diagrams}

\begin{figure}[h!]
  \centering 
   \begin{subfigure}[b]{0.48\textwidth}
        \centering  \includegraphics[width=\linewidth]{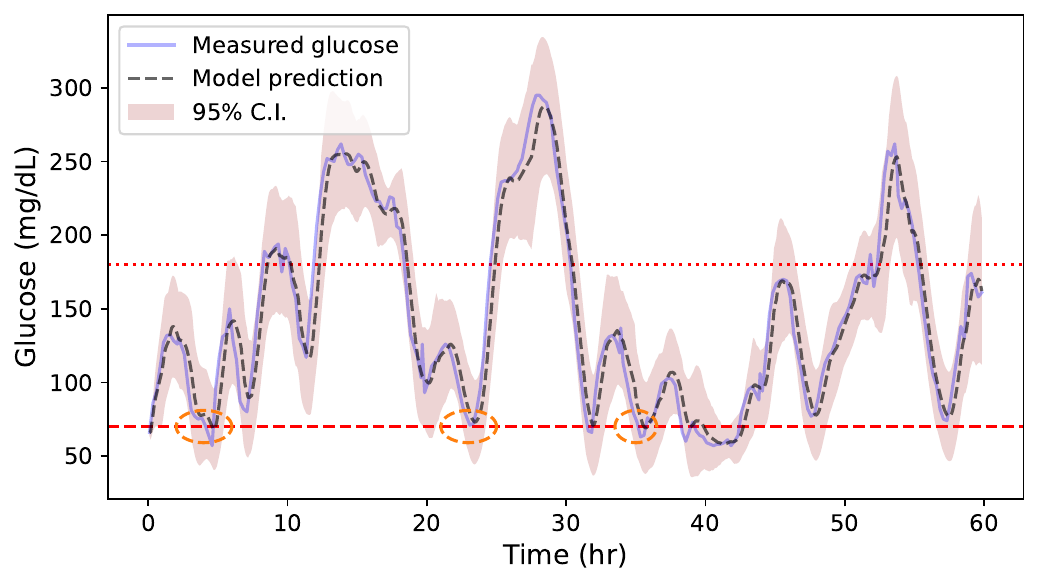}
        \caption{}   \label{fig:sample_plot_evidential}
    \end{subfigure}
    \hfill
     \begin{subfigure}[b]{0.48\textwidth}
        \centering
 \includegraphics[width=\linewidth]{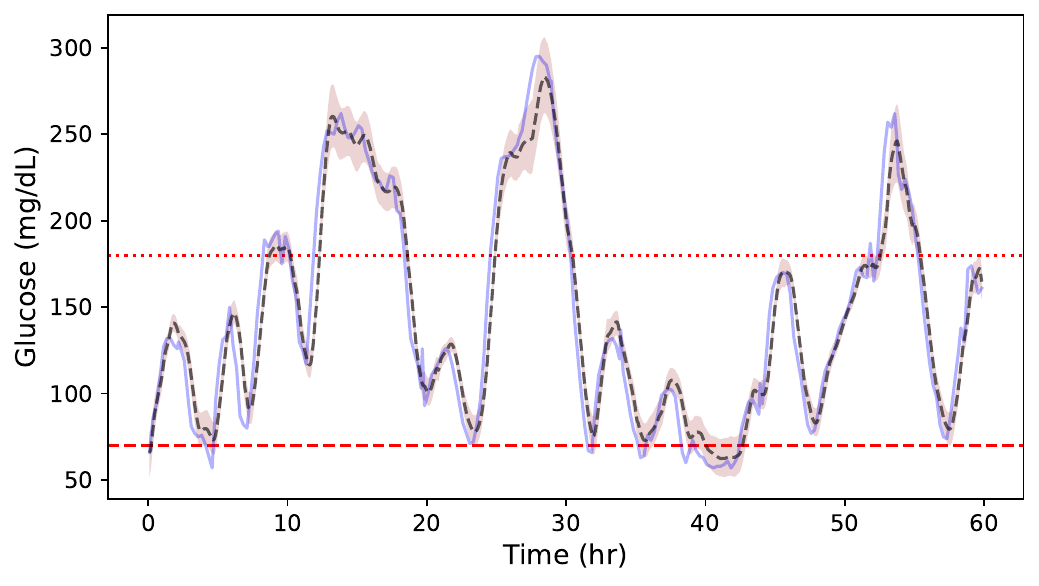}
        \caption{}   \label{fig:sample_plot_drop}
    \end{subfigure}
    \caption{The rolling forecast plot for TEM model (a) and Transformer-based model with Monte Carlo dropout (b) trained on basal insulin-included inputs with 1 hour horizon for Patient with ID `3' in the HUPA dataset \cite{HUPA}, with hypoglycemia ($<70$ mg/dL) and hyperglycemia ($>180$ mg/dL) thresholds being represented by red, horizontal dashed lines. Dashed ellipses locate examples of regions where model would fail to detect hypoglycemia events unless uncertainty estimates are used for detection thresholds.}
    \label{fig:rolling_forecast}
\end{figure}

Figs.~\ref{fig:sample_plot_evidential} and~\ref{fig:sample_plot_drop} are rolling forecast plots -- at each temporal point, we took the mean of all overlapping prediction windows which encompassed the point, and similarly for the $\pm 2\sigma$ confidence bands surrounding it. Fig.~\ref{fig:sample_plot_evidential} shows the full forecast plot for the TEM model trained on basal insulin-included inputs with 1-hr horizon for a specific patient (labeled as `3' in the HUPA-UCM dataset).
The three dashed ellipses locate three examples of temporal regions where model would fail to detect hypoglycemia events unless uncertainty estimates are used as detection thresholds. The $2\sigma$ uncertainty band is visibly larger at inflection points. 
This was not always the case for a generic model across our experiments. Fig.~\ref{fig:sample_plot_drop} plots the rolling forecast for the Transformer model equipped with Monte Carlo dropout trained on the same inputs and patient as Fig.~\ref{fig:sample_plot_evidential}. We can see that there is a much weaker correlation between prediction errors and uncertainty estimates, with a number of regions where the uncertainty bands do not adequately cover the deviations between model predictions and the measured values. This is reflected in the relatively higher MCE, the over-confident empirical coverage probability plot (see Fig.~\ref{fig:ecp}) and the lower Spearman's coefficient $\rho$ between uncertainty estimates and errors. In this case, the uncertainty bands cannot be interpreted robustly and we have a poorly calibrated uncertainty quantification system.

\begin{figure}[h!]
  \centering 
   \begin{subfigure}[b]{0.48\textwidth}
        \centering   \includegraphics[width=\linewidth]{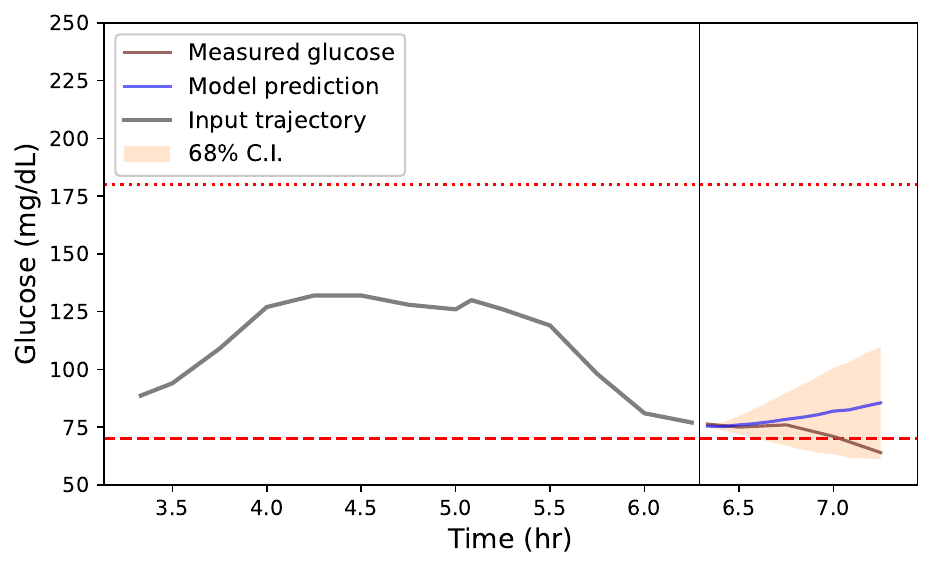}
        \caption{}
        \label{fig:hypo_single}
    \end{subfigure}
    \hfill
     \begin{subfigure}[b]{0.48\textwidth}
        \centering   \includegraphics[width=\linewidth]{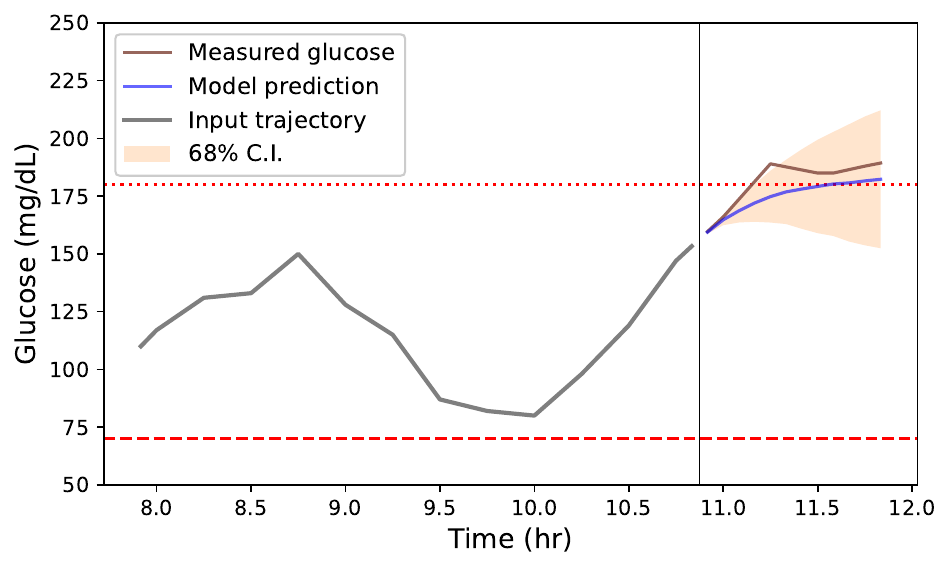}
        \caption{}
        \label{fig:hyper_single}
    \end{subfigure}
    \caption{Single long horizon prediction windows for a couple of points near hypo- and hyperglycemia thresholds with the vertical lines in each diagram marking the start of the prediction horizon.      
    }
    \label{fig:single_hori}
\end{figure}

In Fig.~\ref{fig:single_hori}, we show a couple of single horizon predictions for the TEM model, zooming onto two specific points which are local neighborhoods of hypoglycemia and hyperglycemia events. For example, in Fig.~\ref{fig:hypo_single}, the predicted glucose is deviating away from and above the hypoglycemia threshold even while the actual measured glucose has crossed over, but the $1\sigma$ uncertainty interval encompasses the hypoglycemia region, and thus could have provided appropriate an alert warning if used as detection threshold in the prediction system. Figs.~\ref{fig:rolling_forecast} and~\ref{fig:single_hori} illustrate how uncertainty estimates can be useful for enhancing detection of hypo- and hyperglycemia events even when model predictions deviate from actual glucose trajectories.

\section{Discussion}

Our various experimental results showed evidential regression-based models to be more effective in terms of uncertainty quantification than Monte Carlo dropout and the baseline ridge regression method. This superiority is characterized by lower Brier and higher PR-AUC scores for adverse glycemic event identification, lower MCE and significantly higher Spearman's $\rho$ between uncertainty and errors/clinical risk zones. 
Calibration diagrams showed that for evidential models, the empirical coverage probabilities were much closer to nominal ones relative to other models, with Monte Carlo dropout's uncertainty quantification being over-confident and that of ridge regression being under-confident. In terms of the accuracy metrics, 
evidential models were consistently better than their dropout and ordinary (no uncertainty quantification) counterparts in the aspect of MARD for all input types and both horizon durations. For the longer horizon experiments, in the aspect of DTS zone A accuracy, evidential models were also consistently superior than their counterparts in the longer horizon setting. Overall, these trends indicated that apart from equipping models with well-calibrated uncertainty quantification, evidential models were generally superior in terms of accuracy defined via these clinically relevant metrics.

Within the spectrum of various evidential models trained on the three base network architectures, we identify the TEM model to be most effective in uncertainty quantification as measured by its superior Spearman's $\rho$, $\rho_z$, MCE, the PR-AUC and Brier scores for hypo- and hyperglycemia event identification, relative to LSTM and GRU-based counterparts. In the aspect of accuracy, it yielded the lowest MARD in every case. For hypoglycemia sensitivity score and DTS zone A accuracy, it also yielded the best scores for most experimental settings.

\subsection*{Comments on how our results relate to clinical deployment}

A primary area where glucose forecasting plays a crucial role in diabetes management lies in generating early warnings for hypo- and hyperglycemia. The forecast diagrams in Figs.~\ref{fig:rolling_forecast} and~\ref{fig:single_hori} depict how the confidence bands could enhance sensitivity to adverse glycemic events by virtue of thresholds derived from uncertainty estimates. For example, in Fig.~\ref{fig:hyper_single}, the actual glucose level has crossed the hyperglycemia about 30 min preceding the model-predicted glucose. This discrepancy is however contained within the $1\sigma$ confidence band which, if being invoked in the early warning system as a threshold, would generate the appropriate signal even when model prediction has appeared to be within the normal glycemic range. Such a threshold is not fixed, and in principle can be determined by setting a required sensitivity level that is accompanied by an acceptable false positive rate. In this aspect, our tabulated values of PR-AUC for all our experiments indicate that the TEM model yielded the best precision-sensitivity tradeoff for both hypoglycemia and hyperglycemia, as is visibly evident in the PR curves in Fig.~\ref{fig:pr_curves}.  
These results collectively indicate TEM to be a plausibly excellent candidate for a real-world probabilistic hypo/hyperglycemia detection model. 

Since the model prediction could be used to motivate significant changes in insulin dose or urgent dietary changes \citep{diabetes} in anticipation of predicted hypoglycemia, clinical deployment of the glucose prediction model requires its users to be alerted to potentially consequential model errors in the ideal scenario. Uncertainty estimates are thus vital towards alerting the human user to model predictions that are accompanied by large uncertainty bands. Ideally, we would want uncertainty estimates that are strongly correlated with clinical risks.
We had thus used $\rho_z$ to measure the relationship between each model's uncertainty estimates and the clinical risk zones defined by an international consensus of diabetes clinicians that leverage their domain knowledge. The TEM models exhibited the highest $\rho, \rho_z$ among all the models we considered, and can be interpreted as the most informative with regards to the clinical risks carried by its predictions. 

Finally, we would like to point out a couple of natural future directions of our work here. 
We have used population-level (all 25 patients of the HUPA dataset \citep{HUPA}) training data in all our experiments. It would be interesting to carry out similar investigations using individual patient's data instead to probe the feasibility and effectiveness of a more personalized glucose prediction model. In \citet{Esha}, the HUPA-UCM dataset was also studied and the authors reported that models trained on individual patients appeared to have similar prediction accuracy as those trained at a population level. There are many other input feature combinations for which we can perform a comparative study. Here, we have kept to four-feature inputs similar to \citet{Esha}, with glucose, bolus insulin and carbohydrates being included in all four classes of inputs examined. It would be worthwhile to consider more extensive permutations of input variables.

\section{Conclusion}

In this work, we have examined a number of uncertainty-aware neural network models for blood glucose prediction and adverse glycemic event identification relevant for Type 1 diabetes management. With the base models being one of the three families of sequence models defined by LSTM, GRU, and Transformer architectures, uncertainty quantification was implemented by either Monte Carlo dropout or through evidential output layers compatible with Deep Evidential Regression. Using the HUPA-UCM diabetes dataset of \cite{HUPA} for validation, we identified Transformer-based models equipped with evidential outputs to furnish the most effective uncertainty-aware framework, achieving consistently higher predictive accuracies as measured by MARD and DTS error grid zone A accuracy, as well as better-calibrated uncertainty estimates whose magnitudes significantly correlated with prediction errors and clinical risk zones. This class of models deserves further study as a promising candidate for an uncertainty-aware glucose prediction system that can complement traditional adverse glycemic event prediction techniques. Overall, our work has shown that incorporating uncertainty quantification through evidential regression can lead to robust glucose prediction models that are not merely accurate, but also rigorously informative with regards to its clinical risks as defined by an international consensus of glucose monitoring experts in \citet{DTS}.



\bibliography{BG_pred}

\appendix

\section{Technical details of evidential regression}\label{apd:evidential}

As evidential regression is less commonly used in uncertainty quantification compared to the more conventional Monte Carlo dropout technique \citep{dropout}, and we have also used a relatively novel regularizer first proposed in \citet{tan2025} (in a different context), here we furnish a detailed explanation on how to incorporate evidential regression of \citet{amini} into various time-series models.  

In the following, we explain 
the operational principles of Deep Evidential Regression \citep{amini,sensoy} and how we incorporate it as an uncertainty quantification technique into our models. 

Let $\mathcal{M}(z)$ denote the base neural network with $z$ being its input. Here we consider the architectures of 
long short-term memory (LSTM), gated recurrent units (GRU) and Transformers for $\mathcal{M}(z)$. The uncertainty estimates for each output neuron can be obtained as follows. We first consider a probabilistic model where each output neuron is accompanied by another one representing its uncertainty. This pair of neurons can be interpreted as the Gaussian mean and variance for the probabilistic target. We further assume prior distributions for the mean $\mu$ and variance $\sigma^2$ and integrate $(\mu,\sigma^2)$ out to obtain a marginal distribution that depends on the observed data and the parameters of the prior distribution. Specifically asserting the prior to be a normal-inverse-gamma distribution (NIG) with $\mu \sim \mathcal{N}(\gamma, \sigma^2/\nu)$, $\sigma^2 \sim \Gamma^{-1}(\alpha, \beta)$, we derive the marginal distribution to be
a t-distribution as follows. 
\begin{align}
\label{t_distribution}
&\iint d\mu d\sigma^2\,\,\,
f_{\mathcal{N}} (y_{obs};\mu, \sigma^2)
f_{NIG}(\mu, \sigma^2; \alpha, \beta, \nu, \gamma) \cr
&= 
\frac{\Gamma\left( \alpha + \frac{1}{2} \right)}{\Gamma \left( \alpha  \right)  \sqrt{2\pi \beta  (1+\nu)/\nu  }}
\left(
1 + \frac{(y_{obs} - \gamma)^2}{2\beta  (1+\nu)/\nu}
\right)^{-(\alpha+\frac{1}{2})},
\end{align}
where $f_{\mathcal{N}}$ denotes the auxiliary Gaussian distribution, $f_{NIG}$ the NIG prior and $y_{obs}$ the observed data. In \citet{amini}, the choice of these priors was not quite explicitly explained, although the NIG distribution is known to be a rather standard conjugate prior for Gaussian with unknown mean and variance \citep{gelman}. An advantage of using the NIG prior is that the double integral in eqn.~\eqref{t_distribution} can be analytically done and we have an explicit form for the marginal distribution of which negative logarithm is the loss function. 

Thus, instead of just a single output neuron or a pair representing ($\mu, \sigma^2$), the model $\mathcal{M}(z)$ now has four output neurons ($\alpha, \beta, \nu, \gamma$)
where $\gamma$ represents the mean and $\alpha, \beta, \nu$ are related to the predictive variance $\sigma^2_p$ as follows. 
\begin{equation}
\label{uncertainties}
\sigma^2_a = \mathbb{E}(\sigma^2) = \frac{\beta}{\alpha - 1},\,\,
\sigma^2_e = \text{Var}(\mu)
 = \frac{\beta}{(\alpha - 1)\nu}, \,\, \sigma^2_p = \sigma^2_a + \sigma^2_e,
\end{equation}
where $\sigma^2_a, \sigma^2_e$
denote the aleatoric and epistemic uncertainties respectively. $\sigma^2_a$ is typically interpreted as the uncertainty related to measurement noise while $\sigma^2_e$ represents the uncertainty due to data insufficiency and the underlying model's capacity to represent the observed knowledge \citep{kendall}. 
$\sigma_p^2$ is the variance of the t-distribution defined in the RHS of eqn.~\eqref{t_distribution}.
The model $\mathcal{M}(z)$ yields predictions together with the overall uncertainty expressed by $\sigma^2_p$ which is a simple function of the four output neurons
\be
\sigma^2_p = \left( 1 + \frac{1}{\nu} \right) \frac{\beta}{\alpha - 1}.
\ee
It is trained using a loss function consistent with eqn.~\eqref{t_distribution}, and with $\alpha>1, \nu >0, \beta>0$. Following \citet{amini}, we take the negative log-likelihood of eqn.~\eqref{t_distribution} as a generalized data loss term 
$\mathcal{L}_{data}$
\begin{equation}
\label{dataloss}
\mathcal{L}_{data} = - \log \left[
\frac{\Gamma\left( \alpha + \frac{1}{2} \right)
\left(
1 + \frac{(y_{obs} - \gamma)^2}{2\beta  (1+\nu)/\nu}
\right)^{-(\alpha+\frac{1}{2})}
}{\Gamma \left( \alpha  \right)  \sqrt{2\pi \beta  (1+\nu)/\nu  }}
\right].
\end{equation}
This replaces the usual mean-squared-error loss term, with the original output layer expanded to a tuple of 4 neurons representing $\{\alpha, \beta, \nu, \gamma \}$ per target variable. In terms of model structure (e.g. number of layers, etc.), within this minimal setup, we note that no other
modifications are required for the base model $\mathcal{M}(z)$ apart from expanding the output layer to one that replaces each neuron by a tuple of 4 neurons. The caveat is that we have to train the model using a much more complicated loss function expressed in eqn.~\eqref{dataloss}.

For our glucose prediction models, the inputs $z$ are multivariate physiological signals over a sliding window of some fixed period. In this work, we take the inputs to consist of four-feature sequences over a sliding window of 3 hours. The data is sampled at 5 minutes interval so the input sequence is of length 36. Symbolically,
\begin{equation}
z_t = \left( 
x_{t-35}, \ldots, 
x_t
\right) \in \mathbb{R}^{36\times 4}.
\end{equation}
The model $\mathcal{M}(z)$ predicts the future glucose trajectory over some fixed horizon $h_L$. Denoting the set of predicted glucose by $\hat{y}$,
\begin{equation}
\label{pred_g}
\hat{y} = \left(
g_{t+1}, 
\ldots,
g_{t+h_L} \right)
\in \mathbb{R}^{h_L},
\end{equation}
where $g_t$ denotes the predicted glucose level at time $t$.
In our work, we consider two specific values of $h_L = \{6, 12\}$ representing a 30 minutes and 1 hour forecast horizon respectively.\footnote{These values were also adopted by \citet{Zhu}.}
Minimally incorporating evidential regression within the model translates to expanding the output layer such that apart from eqn.~\eqref{pred_g}, we have 
\begin{align}
\hat{\gamma} &= \left(
g_{t+1}, 
\ldots,
g_{t+h_L} \right), 
\hat{\alpha} = \left(
\alpha_{t+1}, 
\ldots,
\alpha_{t+h_L} \right), \cr
\hat{\beta} &= \left(
\beta_{t+1}, 
\ldots,
\beta_{t+h_L} \right), 
\hat{\nu} = \left(
\nu_{t+1}, 
\ldots,
\nu_{t+h_L} \right),
\end{align}
where $\hat{\gamma}$ denotes the {\itshape mean} glucose level over the prediction horizon, and $\hat{\alpha}, \hat{\beta}, \hat{\nu}$ are the uncertainty-related outputs that can be used following eqn.~\eqref{uncertainties} to construct statistical confidence intervals throughout the prediction horizon. To summarize, 
the evidential regression models learn a mapping
$
\mathcal{M}: \mathbb{R}^{36\times 4} 
\rightarrow \mathbb{R}^{4 \times h_L}$, where, for each forecast horizon step, the network outputs the four parameters of the evidential predictive distribution.

\subsection*{A regularization term for the loss function}
A subtle aspect of evidential regression lies in the regularization of the learning of uncertainties via augmenting $\mathcal{L}_{data}$ with another loss term designed such that ideally, larger uncertainties are associated with greater prediction errors during model training. 
In \cite{amini}, the authors proposed a regularizer constructed to minimize evidence on incorrect predictions that reads 
$\mathcal{L}_{R} = |y_{obs} - \gamma|(2\nu + \alpha)$. Although it was claimed in \cite{amini} to be reasonable, it is unfortunately independent of $\beta$ which controls the overall uncertainty scale, as shown in eqn.~\eqref{uncertainties}. In \citet{tan2025}, the authors proposed a refinement of this original EDL regularizer by multiplying it to the Kullback-Leibler divergence \citep{thomas,Kullback} between the distribution described by eqn.~\eqref{t_distribution} and a weakly-informative prior, its final form being
\begin{equation}
\label{kl}
\mathcal{L}_{kl} = |y_{obs} - \gamma| \left(
\log \frac{\beta^\alpha_r \, \Gamma(\alpha)} {\beta^\alpha} + \gamma_e (\alpha - 1)
+ \frac{\beta- \beta_r}{\beta_r}
\right),
\end{equation}
where $\gamma_e$ is the Euler-Mascheroni constant. In eqn.~\eqref{kl}, 
$\beta_r$ is a reference $\beta$ value which is understood to characterize a weakly informative prior for larger deviations from the observed dataset. We set it to be of the same order-of-magnitude as the maximal value in our glucose dataset. 
The authors of \cite{tan2025} found this regularizer to perform better than $\mathcal{L}_R$ in the context of Physics-Informed Neural Networks (equipped with evidential regression). Theoretically, it aligns with the goal of guiding distributions for erroneous predictions to inherit weak priors with higher uncertainties. To our knowledge, no previous work has considered this information-theoretic regularization term in regression-type contexts. 

In the following, we derive
the regularization term 
in eqn.~\eqref{kl} as first presented in \citet{tan2025}. 
It is mainly derived from considering the KL divergence between the normal-inverse-gamma (NIG) distribution parametrized by $\{\gamma, \alpha, \beta, \nu \}$
and another reference NIG distribution 
with $\{\gamma_r, \alpha_r, \beta_r, \nu_r \}$ that presumably characterizes a non-/weakly informative prior.  
By definition, the KL divergence between two NIG distributions with density functions $(\tilde{p}_{r}, \tilde{p})$ is
\begin{equation}
\label{kl_deriv}
D_{KL}\left( \tilde{p}_{r} \vert| \tilde{p}\right) = \iint \,d\overline{u} \,d\sigma^2_u \,\, \tilde{p}_{r} \log
\frac{\tilde{p}_{r}}{\tilde{p}},
\end{equation}
where $(\tilde{p}_{r}, \tilde{p})$ are priors for a normal distribution with mean
$\overline{u}$ and variance $\sigma^2_u$ which are the integration variables in eqn.~\eqref{kl_deriv}. Explicitly, the prior density function reads
\begin{align}
\label{prior_form}
\tilde{p} &\sim 
\sqrt{\frac{\nu}{2\pi \sigma^2_u}}
\frac{\beta^\alpha}{\Gamma (\alpha)} \left(
\frac{1}{\sigma^2_u}
\right)^{\alpha + 1}
\text{exp}\left[
-\frac{2\beta + \nu (\overline{u} - \gamma)^2}{2\sigma^2_u}
\right], \\
\label{prior_form_2}
\tilde{p}_r &\sim 
\sqrt{\frac{\nu_r}{2\pi \sigma^2_u}}
\frac{{\beta_r}^{\alpha_r}}{\Gamma (\alpha_r)} \left(
\frac{1}{\sigma^2_u}
\right)^{\alpha_r + 1}
\text{exp}\left[
-\frac{2\beta_r + \nu_r (\overline{u} - \gamma_r)^2}{2\sigma^2_u}
\right].
\end{align}
We would like the reference prior to be a non- or weakly-informative prior. Taking the limit of $\nu_r \rightarrow \infty$ in eqn.~\eqref{prior_form_2} yields a uniform distribution over $\mathbb{R}$ which is unfortunately unnormalizable. To avoid this pathology, we take the reference NIG to be parametrized by values of $\gamma, \nu$ identical to those of the evidential model outputs' distributions, and instead set the reference inverse-gamma distribution to be of relatively higher entropy, taking $\alpha_r =1$ and letting $\beta_r$ be task-dependent. Since the variance $\sigma^2_u$
is not physically expected to be larger than the upper bound for $u$, a natural candidate for $\beta_r$ is the maximal glucose value in our training dataset. 
Using this definition of $\tilde{p}_r$ in the double integral of eqn.~\eqref{kl_deriv} leads to the KL divergence term being 
\begin{equation}
\label{div_kl}
\mathcal{D}_{KL}=
\left(
\log \frac{\beta^\alpha_r \, \Gamma(\alpha)} {\beta^\alpha} + \gamma_e (\alpha - 1)
+ \frac{\beta- \beta_r}{\beta_r}
\right).
\end{equation}
Further, like in \cite{amini}, we consider using a simple linear term $|y_{obs} -\gamma|$ that aims to increase the loss for predictions with larger errors, and multiply it to the KL term in eqn.~\eqref{div_kl} to obtain our final expression for the regularizer loss term
\begin{equation}
\mathcal{L}_{kl} = |y_{obs} - \gamma| D_{kl}.
\end{equation}


\section{Some visual plots}\label{apd:plots}

\subsection{Precision-Recall curves}
\begin{figure}[h!]
  \centering \includegraphics[width=0.9\linewidth]{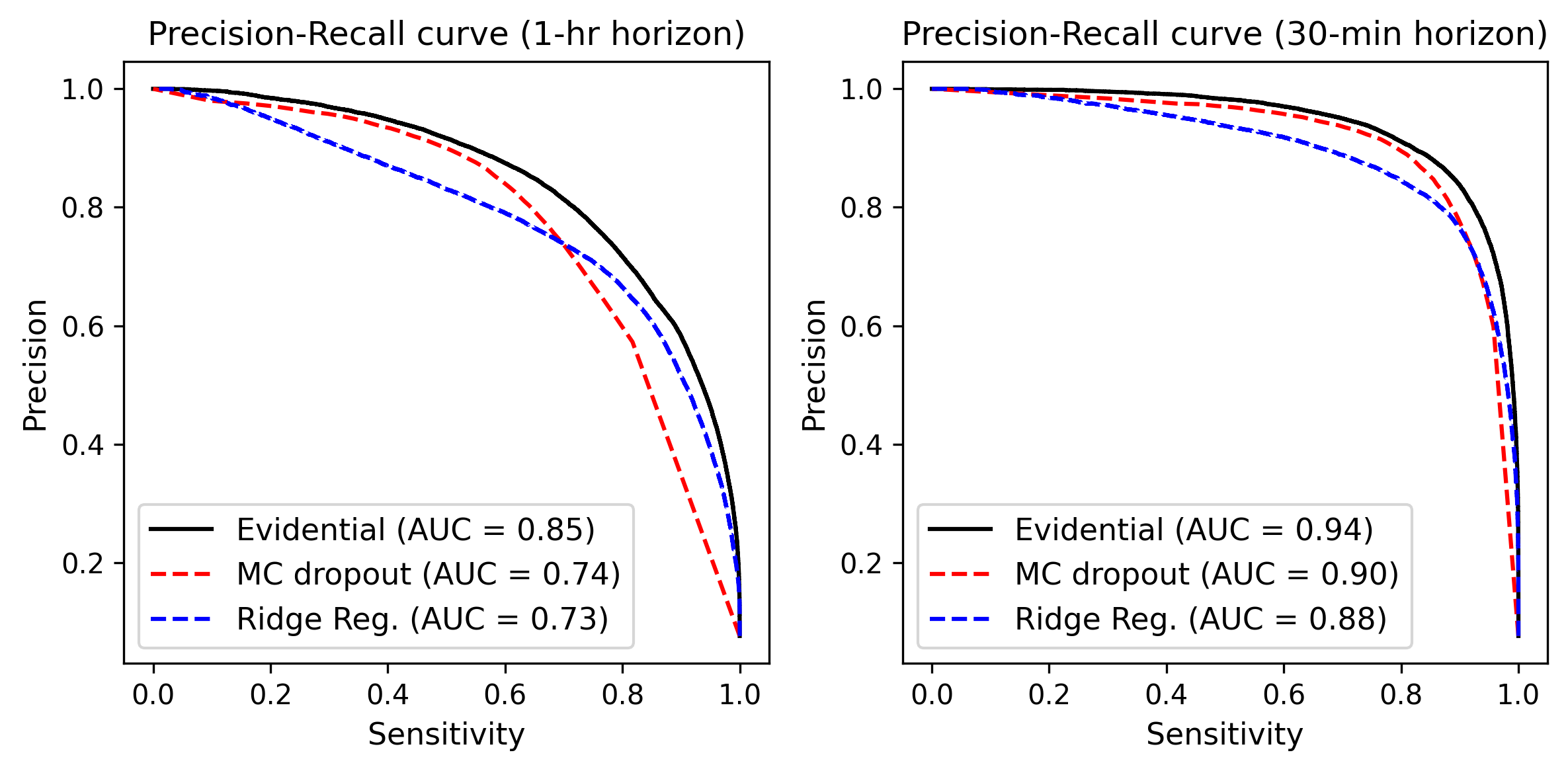}
\caption{Precision-Recall curves for hypoglycemia detection performed by various models trained on heart rate-included inputs. Similar relative trends are also observed for the hyperglycemia case - see AUC values collected in Tables~\ref{tab:heart_short} - \ref{tab:heart_long}. }
\label{fig:pr_curves}
\end{figure}

In Fig.~\ref{fig:pr_curves}, we
display the precision-recall curves for hypoglycemia detection by Transformer Evidential model (TEM) trained on heart rate-included inputs, together with those of the baseline ridge regression and Monte Carlo dropout methods for comparison. One can see that the TEM model consistently achieves higher precision than the other two at every recall level. Its precision–recall trade-off is more visibly superior in the longer predictive horizon setting, as also reflected in the AUC values.

\subsection{On DTS error grids
}
\begin{figure}[h!]
  \centering \includegraphics[width=0.8\linewidth]{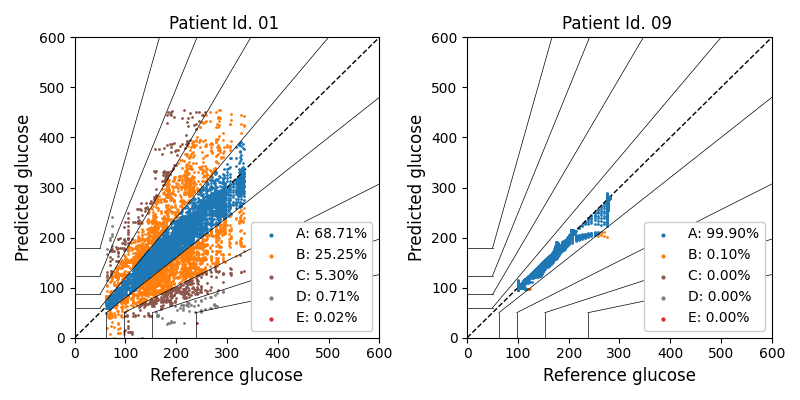}
\caption{DTS error grids for two patients with marked difference in their risk zone distributions. The various pairs of solid lines defined away from the dashed diagonal mark the boundaries of the five risk zones.  }
\label{fig:DTS_sample}
\end{figure}

In Fig.~\ref{fig:DTS_sample}, 
we display a couple of DTS error grid diagrams for two particular patients with relatively larger differences in their risk zone distributions. The model predictions for Patient `1' of \cite{HUPA} are characterized by 
a larger spread in risk zone levels in contrast to those for Patient `9' of \cite{HUPA}. The DTS error grid diagrams can be used as a visual aid in assessing model reliability which is lower for Patient `1' relative to Patient `9'.

\subsection{Boxplots for visualizing input feature dependence of per-patient-defined metrics}

\begin{figure}[ht]
  \centering \includegraphics[width=\linewidth]{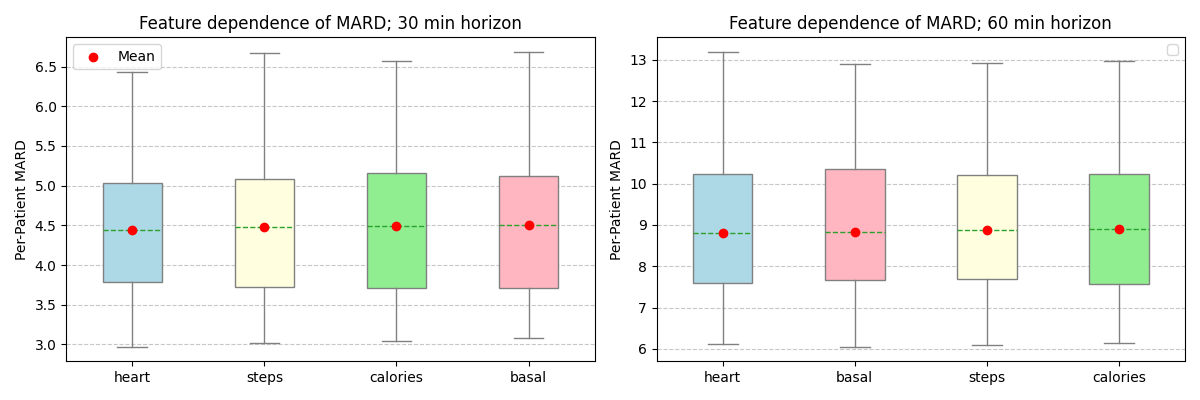}
\caption{Boxplots showing the distribution of per patient MARD values for each class of input features. Friedman test was significant $(p<0.01)$ for the short horizon setting, with post-hoc Wilcoxon signed-rank tests with Holm correction indicating heart rate-trained models to achieve better MARD than those trained on steps and calories (Holm-adjusted $p\sim 0.01$).}
\label{fig:mard_feature}
\end{figure}

\begin{figure}[ht]
  \centering \includegraphics[width=\linewidth]{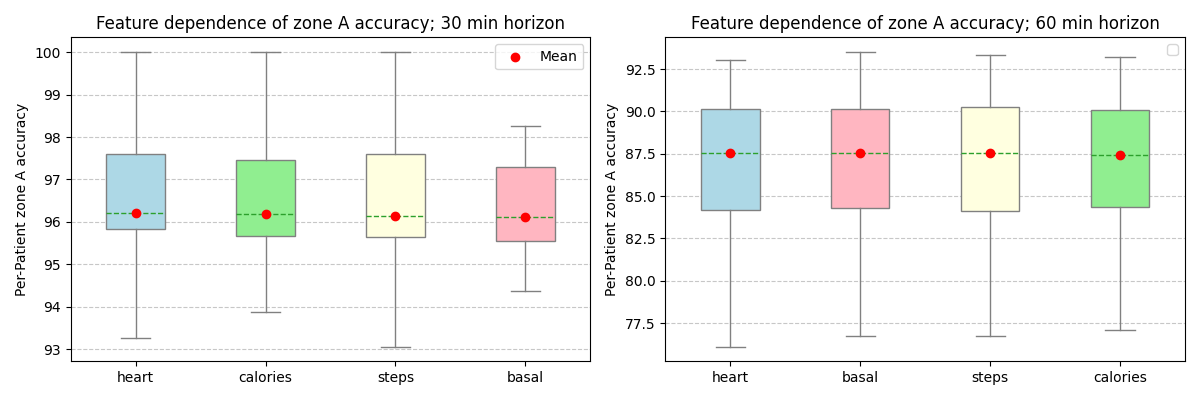}
\caption{Boxplots showing the distribution of per patient DTS zone A accuracy for each class of input features. Friedman test indicated no statistically significant difference among various models.}
\label{fig:zoneA_feature}
\end{figure}

In this work, four families of
4-feature input sequences were considered, each consisting of blood glucose, carbohydrates ingested, bolus insulin and one of the following: heart rate, steps, calories and basal insulin.
Figs.~\ref{fig:mard_feature} and~\ref{fig:zoneA_feature}
display the boxplots for various per-patient-metrics: MARD and zone A accuracy respectively. These accuracy score distributions are those of the TEM model which, at the population level, attained the best accuracy and uncertainty calibration scores. To quantify the degree of sensitivity to the choice of input features, we performed Friedman tests across all four sets of input features and for both choices of horizon durations.  
For MARD values in the 30 min horizon setting, Friedman test revealed a statistically significant effect of input feature choice with statistic $\chi^2(3) = 11.9\,(p< 0.01)$. Post-hoc Wilcoxon signed-rank tests with Holm correction indicated heart rate-trained models with short prediction horizon to achieve better MARD than those trained on steps and calories (Holm-adjusted $p\sim 0.01$). Overall, our analysis did not reveal any significant advantage that any input group has over others.

\section{Tabulation of experimental results related to other input features
}\label{apd:other_tables}

We have performed experiments based on four sets of input features : glucose, bolus insulin, carbohydrates and one of the following $\{$ heart rate, steps, calories and basal insulin $\}$. In the main text, Table~\ref{tab:heart_short} collects various results for heart rate-based inputs in the 30-min horizon setting. In this Appendix, Tables~\ref{tab:basal_short} -  \ref{tab:heart_long} collect various results for models trained on basal insulin, calories and steps-included inputs, as well as the heart rate-included inputs in the 1-hr horizon case.

\begin{table}[ht]
\centering
  \caption{Models trained on basal insulin-included inputs (30 min horizon). Abbreviations: 
zA - zone A accuracy; 
MARD - mean absolute relative difference;
${}^{70}S$ - hypoglycemia sensitivity; ${}^{70}B$ - hypoglycemia Brier score; 
${}^{70}A$ - hypoglycemia PR-AUC; ${}^{180}\{ S, B, A\}$ - the corresponding metrics for hyperglycemia; MCE - mean calibration error; $\rho$ - Spearman's correlation between ucertainty and error; model subscripts {\itshape d, e} - equipped with Monte Carlo dropout and evidential regression respectively; $\rho_z$ - Spearman's correlation between uncertainty and clinical risk zones as defined in \cite{DTS}. Each shaded cell pertains to the best-performing model for each metric column.
}
\vspace{3pt}
  \label{tab:basal_short}
  \begin{tabular}{c | c c c c c c c c c c c c}
    \toprule
    Model & zA & MARD & ${}^{70}S$ & ${}^{70}A$ & ${}^{70}B$ &
    ${}^{180}S$ & ${}^{180}A$ & 
    ${}^{180}B$ & MCE & $\rho$ & $\rho_z$\\
    \hline
    LSTM & 96.7 & 4.46 & 0.79 &  & ${}$ & 0.90 & 
    \\
    $\text{LSTM}_d$ & 96.6 & 4.64 & 0.88 & 0.88 & 0.018 & 0.92 & 0.95 & 0.023 & 0.10 & 0.16 & 0.071
    \\
    $\text{LSTM}_e$ & 96.6 & 4.26 & 0.94 & 0.88 & 0.016 & \cellcolor{gray!20}0.96 & 0.95 & \cellcolor{gray!20}0.020 & 0.04 & \cellcolor{gray!20}0.68 
    & 0.192
    \\ 
    $\text{Transf}$ & \cellcolor{gray!20}96.9 & 4.36 & 0.79 &  & ${}$ & 0.90 &  & ${}$ & ${}$ & ${}$ 
    \\
    $\text{Transf}_d$ & 96.7 & 4.53 & 0.86 & 0.90 & 0.018 & 0.93 & 0.95 & 0.022 & 0.08 & 0.25 & 0.097
    \\
    $\text{Transf}_e$ & 96.8 & \cellcolor{gray!20}4.20 & 0.94 & \cellcolor{gray!20}0.94 & \cellcolor{gray!20}0.015 & \cellcolor{gray!20}0.96 & \cellcolor{gray!20}0.99 & \cellcolor{gray!20}0.020 & \cellcolor{gray!20}0.02 & \cellcolor{gray!20}0.68 & \cellcolor{gray!20}0.205
    \\
     $\text{GRU}$ & 96.8 & 4.38 & 0.78 &  & & 0.91 &  &  & ${}$ & ${}$ 
    \\
    $\text{GRU}_d$ & 96.6 & 4.77 & 0.91 & 0.90 & 0.020 & 0.93 & 0.96 & 0.022 & 0.07 & 0.20 
    & 0.062
    \\
    $\text{GRU}_e$ & 96.5 & 4.29 & 0.94 & 0.93 & 0.016 & \cellcolor{gray!20}0.96 & 0.98 & \cellcolor{gray!20}0.020 & 0.05 & 0.65 & 0.203
    \\
     Ridge & 95.5 & 5.67 & \cellcolor{gray!20}0.95 & 0.87 & 0.021 & \cellcolor{gray!20}0.96 & 0.97 & 0.026 & 0.12 & 0.48 & 0.199
    \\
    \bottomrule
  \end{tabular}
\end{table}

\begin{table}
\centering
  \caption{Models trained on basal insulin-included inputs (60 min horizon).
}
  \label{tab:basal_long}
  \begin{tabular}{c | c c c c c c c c c c c c}
    \toprule
    Model & zA & MARD & ${}^{70}S$ & ${}^{70}A$ & ${}^{70}B$ &
    ${}^{180}S$ & ${}^{180}A$ & 
    ${}^{180}B$ & MCE & $\rho$ & $\rho_z$\\
    \hline
    LSTM & 87.9 & 8.96 & 0.63 &  & ${}$ & 0.80 & &&&
    \\
    $\text{LSTM}_d$ & 88.0 & 8.92 & 0.73 & 0.71 & 0.035 & 0.82 & 0.87 & 0.046 & 0.22 & 0.14 &
    0.100
    \\
    $\text{LSTM}_e$ & 88.7 & 8.33 & 0.88 & 0.82 & 0.028 & 0.90 & \cellcolor{gray!20}0.95 & 0.038 & 0.02 & 0.67 &
    0.322
    \\ 
    $\text{Transf}$ & 88.5 & 8.63 & 0.62 &  & ${}$ & 0.80 &  & ${}$ & ${}$ & ${}$ 
    \\
    $\text{Transf}_d$ & 88.4 & 8.70 & 0.72 & 0.73 & 0.033 & 0.83 & 0.88 & 0.044 & 0.20 & 0.27 & 0.180
    \\
    $\text{Transf}_e$ & \cellcolor{gray!20}89.2 & \cellcolor{gray!20}8.14 & \cellcolor{gray!20}0.90 & \cellcolor{gray!20}0.84 & \cellcolor{gray!20}0.027 & 0.90 & \cellcolor{gray!20}0.95 & \cellcolor{gray!20}0.037 & \cellcolor{gray!20}0.004 & \cellcolor{gray!20}0.68 & \cellcolor{gray!20}0.334
    \\
     $\text{GRU}$ & 88.3 & 8.71 & 0.62 &  & & 0.80 & &  & ${}$ & ${}$ 
    \\
    $\text{GRU}_d$ & 87.9 & 9.00 & 0.75 & 0.73 & 0.034 & 0.84 & 0.89 & 0.043 & 0.15 & 0.16 
    & 0.063
    \\
    $\text{GRU}_e$ & 88.8 & 8.31 & 0.88 & 0.83 & \cellcolor{gray!20}0.027 & 0.90 & \cellcolor{gray!20}0.95 & 0.038 & 0.01 & 0.66 &
    0.327
    \\
     Ridge & 85.9 & 10.06 & 0.89 & 0.73 & 0.033 & \cellcolor{gray!20}0.92 & 0.92 & 0.044 & 0.10 & 0.52 & 0.314
    \\
    \bottomrule
  \end{tabular}
\end{table}

\begin{table}
\centering
  \caption{Models trained on steps-included inputs (30 min horizon).
}
  \label{tab:steps_short}
  \begin{tabular}{c | c c c c c c c c c c c c}
    \toprule
    Model & zA & MARD & ${}^{70}S$ & ${}^{70}A$ & ${}^{70}B$ &
    ${}^{180}S$ & ${}^{180}A$ & 
    ${}^{180}B$ & MCE & $\rho$ & $\rho_z$\\
    \hline
    LSTM & 96.7 & 4.48 & 0.80 &  & ${}$ & 0.90 & &&
    \\
    $\text{LSTM}_d$ & 96.6 & 4.64 & 0.88 & 0.89 & 0.018 & 0.92 & 0.95 & 0.023 & 0.10 & 0.15 & 0.074
    \\
    $\text{LSTM}_e$ & 96.6 & 4.24 & 0.94 & 0.93 & 0.016 & \cellcolor{gray!20}0.96 & 0.98 & \cellcolor{gray!20}0.020 & 0.04 & \cellcolor{gray!20}0.68 & 0.196
    \\ 
    $\text{Transf}$ & \cellcolor{gray!20}96.9 & 4.38 & 0.79 & & ${}$ & 0.90 &  & ${}$ & ${}$ & ${}$ 
    \\
    $\text{Transf}_d$ & 96.7 & 4.52 & 0.87 & 0.90 & 0.018 & 0.93 & 0.95 & 0.022 & 0.09 & 0.24 & 0.103
    \\
    $\text{Transf}_e$ & 96.8 & \cellcolor{gray!20}4.18 & 0.94 & \cellcolor{gray!20}0.94 & \cellcolor{gray!20}0.015 & \cellcolor{gray!20}0.96 & \cellcolor{gray!20}0.99 & \cellcolor{gray!20}0.020 & \cellcolor{gray!20}0.02 & 0.67 & \cellcolor{gray!20}0.208
    \\
     $\text{GRU}$ & 96.8 & 4.39 & 0.78 &  & & 0.91 &  &  & ${}$ & ${}$ &&
    \\
    $\text{GRU}_d$ & 96.6 & 4.69 & 0.91 & 0.90 & 0.020 & 0.93 & 0.96 & 0.022 & 0.07 & 0.18 & 0.058
    \\
    $\text{GRU}_e$ & 96.5 & 4.30 & 0.94 & 0.93 & 0.016 &\cellcolor{gray!20} 0.96 & 0.98 & \cellcolor{gray!20}0.020 & 0.04 & 0.66 & \cellcolor{gray!20}0.208
    \\
     Ridge & 95.5 & 5.66 & \cellcolor{gray!20}  0.95 & 0.87 & 0.021 & \cellcolor{gray!20}0.96 & 0.97 & 0.026 & 0.12 & 0.48 & 0.202
    \\
    \bottomrule
  \end{tabular}
\end{table}

\begin{table}
\centering
  \caption{Models trained on steps-included inputs (60 min horizon).
}
  \label{tab:steps_long}
  \begin{tabular}{c | c c c c c c c c c c c}
    \toprule
    Model & zA & MARD & ${}^{70}S$ & ${}^{70}A$ & ${}^{70}B$ &
    ${}^{180}S$ & ${}^{180}A$ & 
    ${}^{180}B$ & MCE & $\rho$ & $\rho_z$\\
    \hline
    LSTM & 87.6 & 9.17 & 0.63 &  & ${}$ & 0.80 & &&&
    \\
    $\text{LSTM}_d$ & 88.1 & 8.92 & 0.73 & 0.72 & 0.034 & 0.82 & 0.87 & 0.046 & 0.21 & 0.14 &
    0.096
    \\
    $\text{LSTM}_e$ & 88.8 & 8.31 & 0.88 & 0.82 & 0.028 & 0.90 & \cellcolor{gray!20}0.95 & 0.038 & 0.02 & 0.67 & 0.326
    \\ 
    $\text{Transf}$ & 88.6 & 8.63 & 0.62 & & ${}$ & 0.80 &  & ${}$ & ${}$ & ${}$ 
    \\
    $\text{Transf}_d$ & 88.5 & 8.70 & 0.71 & 0.74 & 0.033 & 0.83 & 0.88 & 0.044 & 0.20 & 0.31 & 0.180
    \\
    $\text{Transf}_e$ & \cellcolor{gray!20}89.3 & \cellcolor{gray!20}8.12 & \cellcolor{gray!20}0.89 & \cellcolor{gray!20}0.85 & \cellcolor{gray!20}0.026 & 0.90 & \cellcolor{gray!20}0.95 & \cellcolor{gray!20}0.037 & \cellcolor{gray!20}0.007 & \cellcolor{gray!20}0.68 &
    \cellcolor{gray!20}0.338
    \\
     $\text{GRU}$ & 88.6 & 8.60 & 0.62 &  & & 0.80 &  &  & ${}$ & ${}$ &
    \\
    $\text{GRU}_d$ & 88.2 & 8.90 & 0.76 & 0.74 & 0.033 & 0.83 & 0.89 & 0.043 & 0.14 & 0.16 
    & 0.078
    \\
    $\text{GRU}_e$ & 88.8 & 8.28 & 0.87 & 0.84 &  0.027 & 0.90 & \cellcolor{gray!20}0.95 & 0.038 & 0.02 & 0.65 
    & 0.336
    \\
     Ridge & 86.0 & 10.03 & \cellcolor{gray!20} 0.89 & 0.73 & 0.033 & \cellcolor{gray!20}0.92 & 0.92 & 0.044 & 0.10 & 0.52 & 0.316
    \\
    \bottomrule
  \end{tabular}
\end{table}

\begin{table}
\centering
  \caption{Models trained on calories-included inputs (30 min horizon).
}
  \label{tab:calories_short}
  \begin{tabular}{c | c c c c c c c c c c c c}
    \toprule
    Model & zA & MARD & ${}^{70}S$ & ${}^{70}A$ & ${}^{70}B$ &
    ${}^{180}S$ & ${}^{180}A$ & 
    ${}^{180}B$ & MCE & $\rho$ & $\rho_z$\\
    \hline
    LSTM & 96.7 & 4.58 & 0.79 &  & ${}$ & 0.90 & 
    \\
    $\text{LSTM}_d$ 
    & 96.4 & 5.04 & 0.87 & 0.88 & 0.019 & 0.91 & 0.94 & 0.025 & 0.147 & 0.15 & 0.072
    \\
    $\text{LSTM}_e$ & 96.6 & 4.26 & 0.94 & 0.93 & 0.016 & \cellcolor{gray!20}0.96 & 0.98 & \cellcolor{gray!20}0.020 & 0.043 & 0.67 & 0.194
    \\ 
    $\text{Transf}$ & 96.5 & 4.76 & 0.78 &  & ${}$ & 0.90 &  & ${}$ & ${}$ & ${}$ 
    \\
    $\text{Transf}_d$ & 96.4 & 4.85 & 0.87 & 0.89 & 0.019 & 0.92 & 0.95 & 0.024 & 0.12 & 0.24 & 0.101
    \\
    $\text{Transf}_e$ & \cellcolor{gray!20}96.9 & \cellcolor{gray!20}4.19 &  0.94 & \cellcolor{gray!20}0.94 & \cellcolor{gray!20}0.015 & \cellcolor{gray!20}0.96 & \cellcolor{gray!20}0.99 & \cellcolor{gray!20}0.020 & \cellcolor{gray!20}0.015 &\cellcolor{gray!20}0.68 & \cellcolor{gray!20}0.205
    \\
     $\text{GRU}$ & 96.8 & 4.39 & 0.79 &  & & 0.90 &  &  & ${}$ & ${}$ 
    \\
    $\text{GRU}_d$ & 96.7 & 4.77 & 0.91 & 0.90 & 0.020 & 0.92 & 0.96 & 0.022 & 0.069 & 0.16 & 0.056
    \\
    $\text{GRU}_e$ & 96.5 & 4.33 & 0.93 & 0.93 & 0.016 & \cellcolor{gray!20}0.96 & 0.98 & \cellcolor{gray!20}0.020 & 0.034 & 0.63 & 0.189
    \\
 Ridge Reg. & 95.6 & 5.65 & \cellcolor{gray!20}0.95 & 0.87 & 0.021 & \cellcolor{gray!20}0.96 & 0.97 & 0.025 & 0.121 & 0.47 & 0.201
    \\
    \bottomrule
  \end{tabular}
\end{table}

\begin{table}
\centering
  \caption{Models trained on calories-included inputs (60 min horizon).
}
  \label{tab:calories_long}
  \begin{tabular}{c | c c c c c c c c c c c c}
    \toprule
    Model & zA & MARD & ${}^{70}S$ & ${}^{70}A$ & ${}^{70}B$ &
    ${}^{180}S$ & ${}^{180}A$ & 
    ${}^{180}B$ & MCE & $\rho$ & $\rho_z$\\
    \hline
    LSTM & 87.9 & 9.08 & 0.62 &  & ${}$ & 0.80 & 
    \\
    $\text{LSTM}_d$ & 87.7 & 9.18 & 0.72 & 0.71 & 0.035 & 0.81 & 0.86 & 0.047 & 0.232 & 0.122 & 0.092
    \\
    $\text{LSTM}_e$ & 88.8 & 8.33 & 0.88 & 0.82 & 0.028 & 0.90 & \cellcolor{gray!20}0.95 & 0.038 & \cellcolor{gray!20}0.013 & 0.673 & 0.322
    \\ 
    $\text{Transf}$ & 88.6 & 8.70 & 0.64 &  & ${}$ & 0.80 &  & ${}$ & ${}$ & ${}$ 
    \\
    $\text{Transf}_d$ & 88.2 & 8.88 & 0.71 & 0.73 & 0.034 & 0.83 & 0.88 & 0.045 & 0.206 & 0.302 & 0.181
    \\
    $\text{Transf}_e$ & \cellcolor{gray!20}89.1 & \cellcolor{gray!20}8.17 &  0.88 & \cellcolor{gray!20}0.84 & \cellcolor{gray!20}0.026 & 0.90 & \cellcolor{gray!20}0.95 & \cellcolor{gray!20}0.037 &  0.014 &\cellcolor{gray!20}0.682 & \cellcolor{gray!20}0.338
    \\
     $\text{GRU}$ & 88.5 & 8.65 & 0.62 &  & & 0.80 &  &  & ${}$ & ${}$ 
    \\
    $\text{GRU}_d$ & 88.0 & 9.00 & 0.76 & 0.75 & 0.033 & 0.84 & 0.89 & 0.043 & 0.147 & 0.133 & 0.080
    \\
    $\text{GRU}_e$ & 88.7 & 8.38 & 0.88 & 0.83 & 0.028 & 0.90 & \cellcolor{gray!20}0.95 & 0.038 & 0.018 & 0.650 & 0.331
    \\
     Ridge Reg. & 86.0 & 10.03 & \cellcolor{gray!20}0.89 & 0.73 & 0.032 & \cellcolor{gray!20}0.92 & 0.92 & 0.044 & 0.099 & 0.515 & 0.316
    \\
    \bottomrule
  \end{tabular}
\end{table}

\begin{table}
\centering
  \caption{Models trained on heart rate-included inputs (60 min horizon).
}
  \label{tab:heart_long}
  \begin{tabular}{c | c c c c c c c c c c c c}
    \toprule
    Model & zA & MARD & ${}^{70}S$ & ${}^{70}A$ & ${}^{70}B$ &
    ${}^{180}S$ & ${}^{180}A$ & 
    ${}^{180}B$ & MCE & $\rho$ & $\rho_z$\\
    \hline
    LSTM & 87.5 & 9.26 & 0.63 &  & ${}$ & 0.80 & 
    \\
    $\text{LSTM}_d$ & 87.9 & 8.97 & 0.73 & 0.71 & 0.035 & 0.82 & 0.87 & 0.046 & 0.21 & 0.13 & 0.102
    \\
    $\text{LSTM}_e$ & 88.8 & 8.32 & 0.88 & 0.83 & 0.027 & 0.90 & \cellcolor{gray!20}0.95 & 0.038 & 0.02 & \cellcolor{gray!20}0.68 & 0.327
    \\ 
    $\text{Transf}$ & 88.7 & 8.62 & 0.63 &  & ${}$ & 0.80 &  & ${}$ & ${}$ & ${}$ 
    \\
    $\text{Transf}_d$ & 88.4 & 8.77 & 0.72 & 0.73 & 0.033 & 0.83 & 0.88 & 0.044 & 0.19 & 0.29 & 0.167
    \\
    $\text{Transf}_e$ & \cellcolor{gray!20} 89.2 & \cellcolor{gray!20} 8.11 & \cellcolor{gray!20} 0.89 & \cellcolor{gray!20}0.85 & \cellcolor{gray!20} 0.026 & 0.91 & \cellcolor{gray!20}0.95 & \cellcolor{gray!20} 0.037 & \cellcolor{gray!20} 0.01 &\cellcolor{gray!20} 0.68 & \cellcolor{gray!20}0.337
    \\
     $\text{GRU}$ & 88.6 & 8.65 & 0.63 & & & 0.81 &  &  & ${}$ & ${}$ 
    \\
    $\text{GRU}_d$ & 88.1 & 8.94 & 0.76 & 0.74 & 0.034 & 0.84 & 0.89 & 0.043 & 0.14 & 0.13 & 0.075
    \\
    $\text{GRU}_e$ & 88.8 & 8.26 & 0.88 & 0.84 & 0.027 & 0.90 & \cellcolor{gray!20}0.95 & 0.038 & 0.02 & 0.66 & 0.336
    \\
     Ridge Reg. & 85.9 & 10.10 & 0.88 & 0.74 & 0.032 & \cellcolor{gray!20}0.92 & 0.92 & 0.044 & 0.10 & 0.52 & 0.315
    \\
    \bottomrule
  \end{tabular}
\end{table}

\clearpage
\renewcommand{\thesection}{\arabic{section}}  
\setcounter{section}{0}

\end{document}